\documentclass[10pt,twocolumn,letterpaper]{article}

\usepackage{cvpr}
\usepackage{times}
\usepackage{epsfig}
\usepackage{graphicx}
\usepackage{amsmath}
\usepackage{amssymb}

\usepackage[breaklinks=true,bookmarks=false]{hyperref}
\usepackage{booktabs}
\usepackage{multirow}
\cvprfinalcopy 

\def\cvprPaperID{****} 

\begin{document}

\title{Take the Scenic Route:\\ Improving Generalization in Vision-and-Language Navigation}

\author{Felix Yu \hspace{30pt} Zhiwei Deng \hspace{30pt} Karthik Narasimhan \hspace{30pt} Olga Russakovsky\\
\\
Princeton University\\
{\tt\small \{felixy, zhiweid, karthikn, olgarus\}@cs.princeton.edu}
}

\maketitle

\begin{abstract}
   In the Vision-and-Language Navigation (VLN) task, an agent with egocentric vision navigates to a destination given natural language instructions. The act of manually annotating these instructions is timely and expensive, such that many existing approaches automatically generate additional samples to improve agent performance. However, these approaches still have difficulty generalizing their performance to new environments. In this work, we investigate the popular Room-to-Room (R2R) VLN benchmark and discover that what is important is not only the amount of data you synthesize, but also how you do it. We find that shortest path sampling, which is used by both the R2R benchmark and existing augmentation methods, encode biases in the action space of the agent which we dub as \textbf{action priors}. We then show that these action priors offer one explanation toward the poor generalization of existing works. To mitigate such priors, we propose a path sampling method based on random walks to augment the data. By training with this augmentation strategy, our agent is able to generalize better to unknown environments compared to the baseline, significantly improving model performance in the process.
\end{abstract}

\section{Introduction and Related Works}

\begin{figure}[t]
  \centering
  \includegraphics[width=0.50\textwidth]{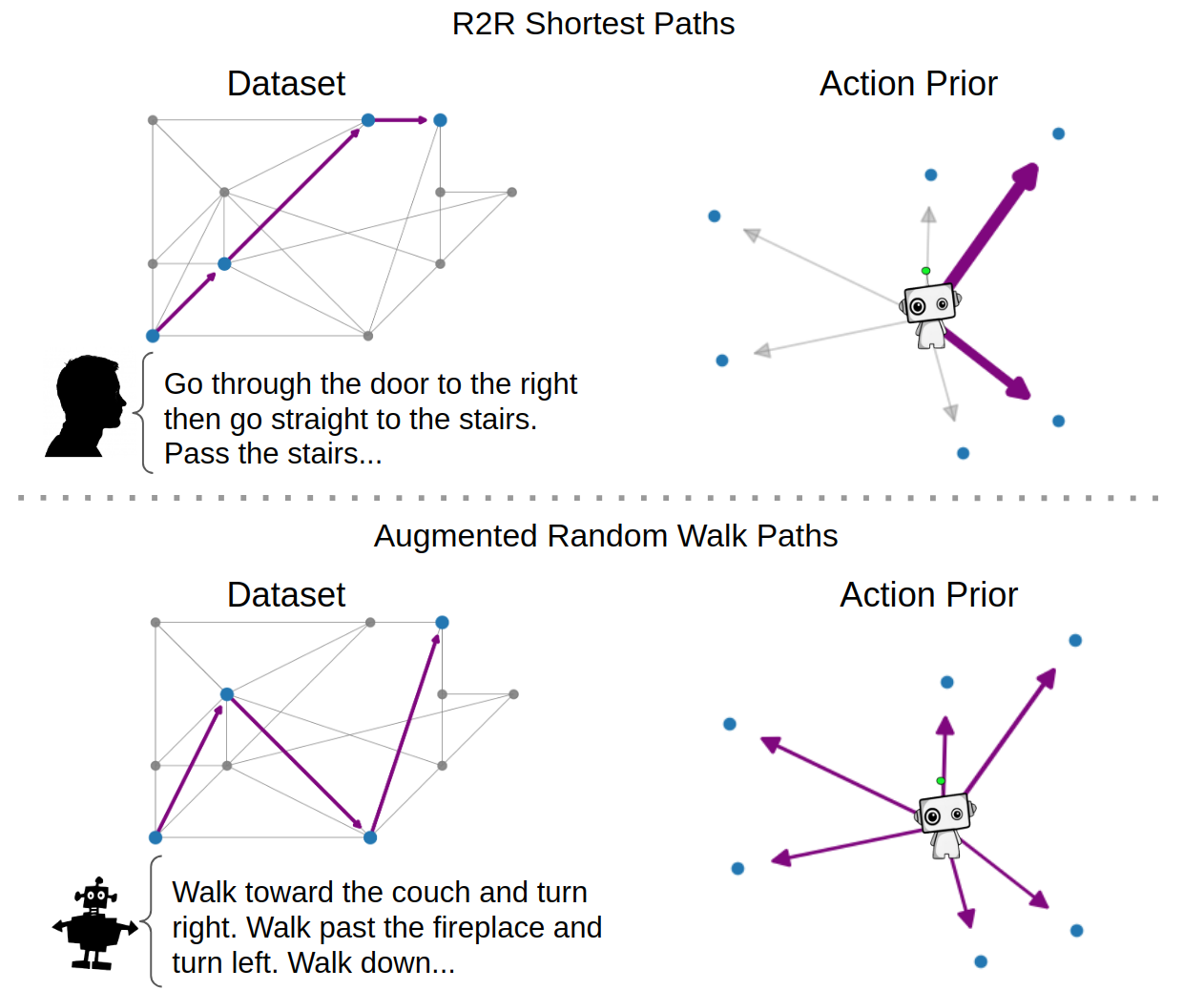}
\caption{The Room-to-Room (R2R) dataset contains a limited number of human annotated instructions for shortest paths. This brings about two problems: Agent performance suffers due to lack of data, and shortest path sampling leads to a skewed distribution in action space, which we refer to as action priors. These action priors can affect agent generalizability to unseen environments. To mitigate both problems, we propose to augment the dataset with additional \textbf{machine} annotated instructions of \textbf{random} walk paths, which do not contain such action priors.}
\label{fig:vln-example}
\end{figure} 

The Vision and Language Navigation (VLN) task is a complex problem which requires a agent to interpret and blend together multiple modalities, including visual scenes and spoken language. In the task, an agent is given a sequence of natural language instructions e.g. "Go through the door to the right then..." and placed at a starting location in the environment. At each timestep, the agent perceives its surrounding visuals through a set of images, each one corresponding to a viewpoint, and performs an action by choosing from a subset of these viewpoints to teleport to (referred to as teleporting action space). A side effect of this complexity is the difficulty in obtaining natural language instructions, which requires annotators to traverse the navigational path through a simulator to write the instruction. As a result, VLN benchmarks such as the Room-to-Room (R2R) task \cite{anderson2018vision} that we focus on in this work only provide a limited amount of annotated data, e.g. 21,567 sets of instructions for R2R. This leads to poor performance and lack of generalization to new environments on the benchmark \cite{anderson2018vision}.

To circumvent this lack of annotations, Fried et. al. \cite{fried2018speaker} propose the \textit{Speaker}, an architecture which is trained on the original R2R dataset to take path trajectories as inputs and output natural language instructions, allowing them to generate an order of magnitude more synthetic instructions from paths sampled through the Matterport3D simulator \cite{chang2017matterport3d, anderson2018vision}. Their work and subsequent works \cite{wang2019reinforced, ma2019self, ke2019tactical, ma2019regretful} which use this augmented data show that performance rises as the amount of augmented data increased, but the generalization gap of these models still exist. 

In this work, we focus not only on improving performance through Speaker-based data augmentation, but also on reducing the generalization gap by changing the type of paths we sample over. We find that existing methods \cite{anderson2018vision, fried2018speaker, wang2019reinforced, ma2019self, ke2019tactical, ma2019regretful, tan2019learning} which use shortest path sampling contains biases over the teleporting action space (dubbed action priors) such that the agent can learn to perform navigation in known environments without relying on the natural language instructions. We hypothesize that since these action priors are specific to each environment, agents are unable to transfer this knowledge to novel scenes, thus leading to the generalization gap. To alleviate these priors, we opt to use random walk path sampling rather than shortest path sampling to augment the existing R2R dataset. By mitigating these scene specific action priors, the agent relies more on cues such as language which generalize better to unseen environments. As a result, we see a significant decrease in the generalization gap from the baseline model, improving performance in unseen environments in the process. 

Other existing works have also tried bridging the generalization gap. Wang et. al. \cite{wang2019reinforced} allows the agent to explore the unseen environments in a self-supervised fashion before evaluating to boost performance. Hu et. al. \cite{hu2019you} proposed an ensemble method using various visual representations, and Tan et. al. \cite{tan2019learning} performs data augmentation on both paths as well as environments by performing consistent visual feature masking. However, all works still train their models purely on shortest paths. To our knowledge, we are the first to investigate the role of these action priors and to propose training navigational agents on non-shortest paths for the R2R benchmark. Although Jain et. al. \cite{jain2019stay} propose the Room-for-Room (R4R) task, which creates non-shortest paths by concatenating paths from R2R together, the R4R task is a separate benchmark altogether. In this work, we focus solely on R2R. 

\section{Room-to-Room and Lack of Generalization}
We first elaborate on the Room-to-Room benchmark \cite{anderson2018vision} and establish notation in the process. Afterwards, we examine how action priors exist in the benchmark and how this affects generalization.
\subsection{Room-to-Room Setup}
In the Room-to-Room (R2R) benchmark, an agent is given natural language instruction $\vec x$ for some path consisting of multiple viewpoints $\vec p = (s_1, s_2, ..., s_{n_p})$, where $s_i$ denotes a single viewpoint and ${n_p}$ is the number of viewpoints in the path. At each point in time $t$, the agent observes a panoramic visual of its current state $s_t$, represented by 36 discrete view vectors $\vec v_{s_t} = \{\vec v_{s_t,i}\}_{i=1}^{36}$. The agent also receives $A_{s_t}$, the action space at the current state. Following the panoramic framework introduced in \cite{fried2018speaker}, rather than using primitive actions such as \texttt{TURN-LEFT, FORWARD, STOP}, each action $a \in A_{s_t}$ corresponds to another location/state that the agent can navigate to (as well as \texttt{STOP}). This means that the action space of the agent depends on the current state, and therefore the environment of the agent. The agent must then combine instruction $\vec x$, visual information $\vec v_{s_t}$, and action history $\{(a_{t'}, s_{t'})\}_{t'=1}^{t-1}$ to choose the action $a_t \in A_{s_t}$ that corresponds to the next location to move to. After either $T$ steps, or when the agent chooses to stop, we evaluate how far the agent's current state is from goal $s_{n_p}$.

This action space brings about an alternate interpretation to the task: Each environment can be interpreted as a graph, where nodes are the states that the agent can be located in, and edges between two nodes denotes direct navigability from one state to the other. At each timestep, the action space of an agent at a state is that state's outgoing edges. Thus, the agent is performing a graph traversal with only local knowledge of the graph. 

\subsection{Overfitting due to Action Priors}
\label{subsection:priors}
From this graph traversal interpretation, it can be seen that the decisions an agent makes at state $s_t$ may not only depend on the natural language instruction, but also on the number of times each outgoing edge is traversed within the training data. If agents are able to recognize their current state given visual information, they can be biased toward choosing actions that appear more frequently within the training data while ignoring other sources of information. 

We examine the action priors that can arise from shortest path sampling by looking at the augmented dataset generated by Fried et. al. \cite{fried2018speaker} and used by other subsequent works \cite{ma2019self, ma2019regretful, wang2019reinforced}, which we will refer to as the Speaker-Follower Augmented Dataset. We choose to use this dataset over the original R2R dataset since it contains a more representative number of shortest path samples. To test how useful these shortest path action priors are in the R2R task, we treat each environmental graph as a Markov chain and calculate their Markov transition matrices (MTM) based on the number of times each edge is traversed within the dataset.

We then feed the MTMs to a greedy agent which takes in no language information. For each for each test sample $p = (s_1, ..., s_{n_p})$ within the validation set of seen environments in the R2R dataset, our greedy agent starts at $s_1$ and takes $T=5$ greedy steps, choosing the action $a \in A_{s_t}$ which has the highest probability according to the MTM. We report the success rate as defined in \cite{anderson2018vision}, which measures the fraction of times the agent stop position $s_T$ is within three meters of $s_{l_p}$. We compare this greedy agent with a random agent which takes $T$ random steps, as well as with the Follower navigational agent reported in Fried et. al. \cite{fried2018speaker} which takes language and vision as input. Our findings are reported in table \ref{table:naive-results}.

As can be seen, our greedy agent performs much better than the random agent, increasing success rate by an absolute $23\%$ (relative $192\%$ increase). Furthermore, although the agent receives no language instructions and has no reasoning capabilities, we are able to achieve a success rate that is over half that of the Follower agent. This is a surprising result, since the Follower agent is given the language instructions and has memory over its action history to perform more sophisticated reasoning. 

From this, we can see that action priors are useful in navigating to goals in seen environments, even in the absence of language instruction. Agents which are able to locate themselves in an environment given visual information $\vec v_{s_t}$ can depend of such priors to perform well on scenes in the training data while ignoring language information given. If we mitigate such priors, the agent may learn to rely more on cues such as language which generalize better to unseen environments.
\begin{table}[t]
\centering
\begin{tabular}{c c c c c} \toprule
  {Input Modality} & \multicolumn{2}{c}{MTM} & & {V + L}\\ \cmidrule{2-3} \cmidrule{5-5}
  {} & {Greedy} & \;\;{Random} & &\;{Follower\cite{fried2018speaker}}\\ \midrule
  {Success Rate} & {0.35} & {0.12} & & {0.66} \\ \bottomrule
\end{tabular}
\caption{Success Rate over Val Seen data split for the Greedy and Random agents, which take in the Markov transition matrices (MTM), and the Follower navigation agent, which takes in vision and language information (V + L).}
\label{table:naive-results}
\end{table}

\section{Methods}
\subsection{Random Walk Path Sampler}

\begin{figure}[t]
  \centering
  \includegraphics[width=0.48\textwidth]{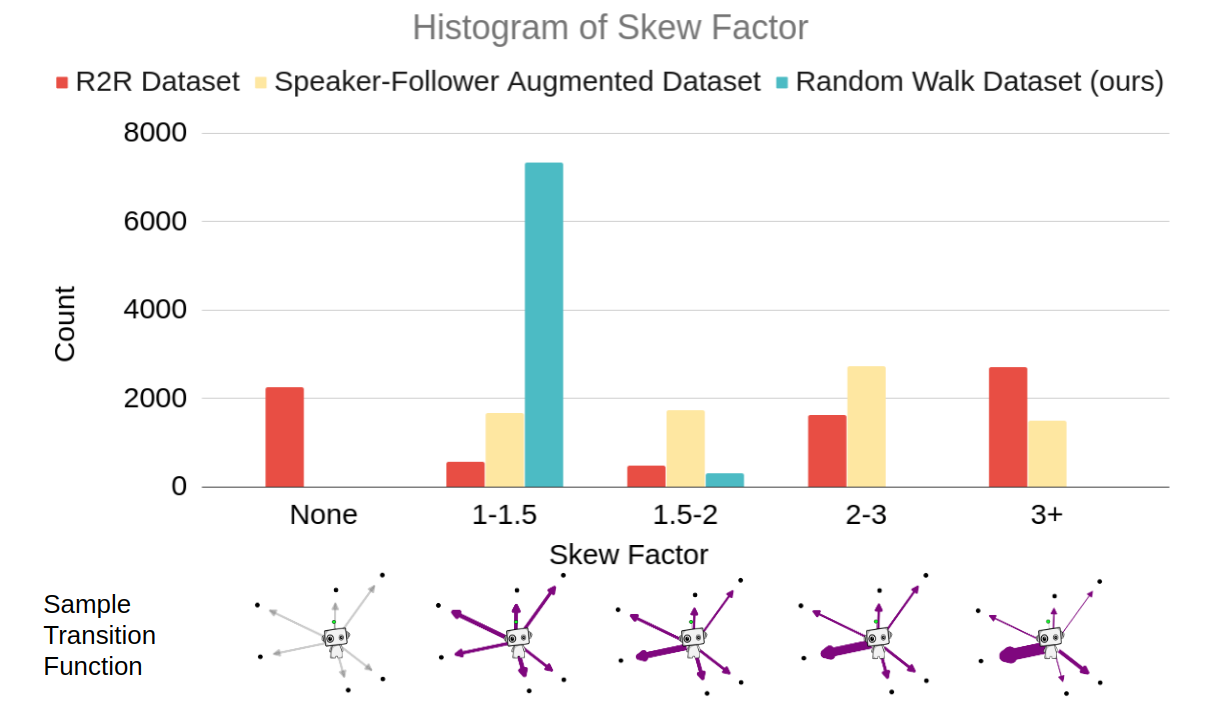}
\caption{Histogram of skew factors: Given the Markov Transition Matrix (MTM) of each environment calculated from the datasets, we calculate the skew factor of each node. This is defined as the ratio between the largest transition probability at that specific node and the probability under uniform distribution. If the node is never visited in the dataset, as can be the case in the R2R dataset, the skew factor is None. We see that although the MTMs for R2R Dataset and Speaker-Follower Augmented Dataset contain large skew factors, almost all skew factors for the Random Walk Sampler MTM has a skew factor close to 1, i.e. the distribution over actions for that node is close to uniform and have minimal action priors.}
\label{fig:skew-hist}
\end{figure}

This motivates us to use random walks rather than shortest paths when augmenting existing R2R dataset. In particular, our sampling is done using the following method: We first uniformly draw a starting viewpoint across all possible viewpoints in the training data. Then, we sample a path length according to the distribution of path lengths found in the R2R benchmark training data. To avoid the actions priors that exist in shortest path sampling, we perform a random walk while avoiding nodes already visited in the path. Finally, if the end goal is not at least three meters away from starting location, we re-sample the path. 

Figure \ref{fig:skew-hist} visualizes the reduction in action priors by using this sampling method over shortest path sampling. We first sample an identical number of paths as the Speaker-Follower Augmented Dataset, and then for each of the original R2R Dataset, Speaker-Follower augmented Dataset, and our Random Walk Dataset, we calculate the Markov transition matrix $\mathbb M$ as done in Section \ref{subsection:priors}. Then, for each node $i$ in the training environments, we calculate it's \textit{skew factor}, which we define as the ratio between the largest transition probability from node $i$ and the probability under uniform distribution. Ideally, we want the skew factor to approach $1$, which denotes that the transition function out of node $i$ is as close to uniform as possible. As can be seen from the histogram, both shortest path datasets contain a significant number of nodes with high skew factors, while the skew factor of almost all (96\%) nodes in our random walk dataset is close to $1$. 

\subsection{Agent Framework}
We now go over the framework used to add random walk paths to the existing R2R dataset to mitigate action priors. Our framework is based off of Speaker-Follower's \cite{fried2018speaker}, and consists of the aforementioned \textit{Path Sampler}, a \textit{Speaker}, and a \textit{Follower} navigational agent, which we will elaborate on shortly. We first pre-train the \textit{Speaker} using the R2R benchmark data and fix the weights. Then, to train the \textit{Follower} on augmented data, we sample random walks with the \textit{Path Sampler} on the fly and annotate instructions using our fixed \textit{Speaker}.

\begin{table*}[t]
\centering

\begin{tabular}{l l l c c c c c c c c c} 
  \toprule
  
  {Condition} & \multirow{2}{2cm}{Data Augmentation} & \multirow{2}{1.5cm}{Forcing Method} & \multicolumn{4}{c}{Seen Validation} & {} & \multicolumn{4}{c}{Unseen Validation} \\ \cmidrule{4-7}\cmidrule{9-12}
   & & &\;{$\downarrow$NE}\; &\;{$\uparrow$SR}\; & \;{$\uparrow$OSR}\; & \;{$\uparrow$SPL}\;& \;\;{}\;\; & \;{$\downarrow$NE}\; & \;{$\uparrow$SR}\; & \;{$\uparrow$OSR}\; & \;{$\uparrow$SPL}\;\\ 
  \midrule 
  & & & \multicolumn{9}{c}{Reported in Speaker-Follower \cite{fried2018speaker}} \\
  \midrule 
  {1} & {None} & {Student} & {4.86} & {52.1} & {63.3} & {-}& {} & {7.07} & {31.2} & {41.3} & {-}\\
  {2} & {Shortest} & {Student} & {3.36} & {66.4} & {73.8} & {-} & {} & {6.62} & {35.5} & {45.0} & {-}\\ 
  \toprule
  & & & \multicolumn{9}{c}{Our Implementation} \\
  \midrule 
  {3} & {None} & {Student} & {4.39} & {57.1} & {68.9} & {47.0} & {} & {6.98} & {27.2} & {38.6} & {18.7}\\
  {4} & {Shortest} & {Student} & {\textbf{3.99}} & {\textbf{61.6}} & {\textbf{69.4}} & {\textbf{54.0}} & {} & {6.85} & {29.7} & {41.0} & {20.1} \\
  {5} & {None} & {Teacher} & {5.36} & {51.6} & {59.4} & {48.6} & {} & {7.13} & {32.5} & {41.1} & {29.0}\\
  {6} & {Shortest} & {Teacher} & {4.97} & {54.0} & {60.5} & {51.7} & {} & {7.12} & {33.8} & {42.2} & {31.0}\\ 
  \midrule
  {7 (ours)} & {Random} & {Teacher} & {5.03} & {53.0} & {61.6} & {50.4} & {} & {\textbf{6.29}} & {\textbf{38.9}} & {\textbf{46.7}} & {\textbf{36.0}}\\ 
  \bottomrule
\end{tabular}
\caption{Reported Results. $\downarrow$ denotes lower is better, $\uparrow$ denotes higher is better. Although using shortest path sampling leads to best overall performance over all metrics on known environments, we see the model trained with our random walk sampling achieves best performance over unseen environments.}
\label{table:main-results}
\end{table*}

\textbf{Speaker} 
The \textit{Speaker} is takes in all visual information $\vec v$ for a path $\vec p = (s_{1}, ..., s_{l_p})$ and generates natural language instructions $\vec x$ according to $p^S(x_t | \vec v, x_{1,...,t-1})$. This is done through a \textsc{Seq2seq} with attention architecture \cite{sutskever2014sequence, luong2015effective}.

\textbf{Follower}
The \textit{Follower} is the navigational agent, and architecturally mirrors the \textit{Speaker}. Given natural language instructions $\vec{x}$ and the environment, the agent defines the navigation distribution as $p^T(a_t|\vec{x}, s_t, \tau)$, where $\tau$ encodes the history of the agent for a particular path traversal. At each timestep $t$, the agent receives visual features $\vec v_{s_t}$ and performs an action $\vec{a}_t \in A_{s_t}$. 


\section{Experiments and Results}

We compare our \textbf{Random Walk} Augmentation Follower trained on both the R2R dataset and augmented random walks against two baseline augmentation methods: (1) \textbf{None}, under which an agent is trained with only the R2R dataset, and (2) \textbf{Shortest}, an agent trained on R2R dataset and augmented shortest paths. All agent architectures remained identical between experiments. We use a batch size of 64, and Adam as the optimizer with a learning rate of 0.0001. All models are trained for 60,000 iterations. If data augmentation is applied, the model is first trained for 40,000 iterations on the augmented data, followed by 20,000 iterations on the original R2R data. 

Following \cite{fried2018speaker}, all models are trained through imitation learning. When all samples are shortest paths, it is possible to re-calculate shortest paths on the fly when models deviate from the original path. This allows us to use \textbf{student} forcing, where the action taken is sampled from the agent's policy at each timestep. Since our random walk sampling method has non-shortest paths, this is not an option and we train that model with only \textbf{teacher} forcing, where the action taken is always the ground truth action. We report metrics for the baselines with both student and teacher forcing. Our models are based off of a re-implementation of \cite{fried2018speaker}, and due to these implementation differences, there are differences in reported metrics. For fairness to both our work and theirs, we report both versions when applicable, but for consistency, run analysis on results gathered from our implementation. 

Our results are given in table \ref{table:main-results}. We evaluate our model primarily using Success Rate (SR) as described in section \ref{subsection:priors}. We also show Navigational Error (NE) which measures distance between goal state and agent's last state, Oracle Success Rate (OSR) which measures success rate at the closet point that the agent ever was to the goal, and Success rate weighted by Path Length (SPL) which normalizes Success Rate by the length of the traversed path.

We can see that for validation samples in seen environments, the training scheme that yields the best results$^{(4)}$ matches that used in \cite{fried2018speaker} with shortest path augmentation and student forcing, with a success rate of 61.6\%. However, we see that this model generalizes poorly to unseen environments with success rate dropping to 29.7\% which is an absolute decrease of 31.9\% and a relative decrease of 51.8\%. On the contrary, our method$^{(7)}$ only sees a performance drop from 53.0\% to 38.9\% giving us absolute decrease of only 14.1\%  and relative decrease of 26.6\%. Furthermore, we can see that although models trained with shortest path augmentation outperform ours trained with random walks, our model outperforms all baselines across all metrics on the unseen environments, improving success rate from the next best model$^{(6)}$ from 33.8\% to 38.9\%. It is helpful to note that we also outperform the original values$^{(2)}$ included in \cite{fried2018speaker}. These results show promise toward our sampling strategy and validate our hypothesis that action priors can negatively impact the generalizability of agents.

\section{Conclusion}
In this paper, we simultaneously deal with the scarcity of data in the R2R task while removing biases in the dataset through random walk data augmentation. By doing so, we are able to reduce the generalization gap and outperform baselines in navigating unknown environments. 

\section{Acknowledgments}
This work is supported by the Princeton CSML DataX fund. We would also like to thank Zeyu Wang, Angelina Wang, and Deniz Oktay for offering insights and comments which guided the paper.
{\small
\bibliographystyle{ieee_fullname}
\bibliography{scenic_route}
}

\end{document}